# Designing an AI-Based Adaptive Controller Augmented with a System Identifier for a Micro-Class Robot Equipped with a Vibrating Actuator

*AmiReza BabaAhmadi

Affiliation: School of Mechanical Engineering, College of Engineering, University of Tehran, Tehran, Iran

Babaahmadi.amir@ut.ac.ir

Nima Naseri

Affiliation: Department of Mechanical Engineering, Sharif University of Technology, Tehran, Iran

## Abstract

In this paper, an adaptive control scheme based on using neural networks is designed to guarantee the desired behavior of a micro-robot which is equipped with vibrating actuators and follows the principle of slip-stick movement. There are two tiny shaking motors which have been utilized to run the micro-class robotic system. Dynamic modeling equations are expressed by considering the spring coefficient of the bases. After that, the effect of the spring on the foundations was investigated. In addition to designing neural-based controller, an AI-based system identifier has been developed to help the controller update its parameters and achieve its desired targets. Using this method, several specific paths for the movement of this micro robot are simulated. Based on the simulation results, the proposed controlling strategy guarantees acceptable performance for tracking different paths due to plotted near-zero errors and handles the nonlinear behavior of the micro-robot system.

## 1- Introduction

Microrobots are widely used in a variety of applications due to their remarkable capabilities such as precision, reliability, and superior performance when compared to human abilities. in [1], a simulation of an ABB IRB120 industrial robot has been performed, and several methods for deriving linear and nonlinear equations for this robot has been discussed. Another publication examined the application of reinforcement learning to a simple multi-body system [2]. However, these robots were lacking enough precision and accuracy in their operational environment. Nowadays, the field of microrobotics has garnered significantly more interest from researchers and roboticists, and micro-robotics applications have experienced a dramatic increase in different industries.

Due to their small size and precision, they have been used in low-volume applications, military applications, hazardous and poisonous areas, medicine, and most notably in drug delivery fields. Robots in micro-class could be characterized based on their performing task and variation in their size [3]. In order to define micro-robots based on their tasks, the "micro-robot" term refers to a class of robots performing a specific task at the micro-scale level.

Designing the systems mentioned above requires designing a system controller to modify manipulation strategies and its operations to guarantee the desired behavior of the system. The functions of microrobots have been extensively discussed in [4]. Micro-robot definitions could be defined according to their general size. According to a valid classification that prominent researchers have defined, the size of a micro-robot varies from 1 in$^3$ down to 100 μm$^3$. Micro-robots which have been studied in this article uses a slip-stick locomotion property for their motion. This feature is established on the stepping-mode mechanism. The consequent vibrations allow the symmetrical slip of the robot's base to move reciprocally. Therefore, the net movement of the micro-class robot would be zero. Provided that the time of vibration, contact forces, and consequently frictional forces of the slip are higher or lower than the other side, the slip value would also change in those two orientations, resulting in a net motion of zero. In this way, slip generation is derived from lateral vibrations, and slip changes are because of contact forces alteration.

Only a few papers published in the field of microrobotics have examined the slip-stick principles of locomotion. Vartholomoes and Papadopoulos introduced a new mini-class robotic platform in [5] based on the stick-slip principle. The micro-robot mentioned above used three tiny-shaking motors to control the total DOF for the system. In [6], they studied the speed control of the new micro-robot platform run by vibrational motors. Moreover, in [7], they examined the simulated performance of the mentioned system equipped with three vibrational micro-actuators.

In [8], a brief description of how to assemble hardware and the process of the new platform has been proposed. In [9], a new microrobot named Kilobot was introduced for the first time using the stick-slip movement principle. In [10], the speed control of micro-motors considering their vibrational behavior has been investigated.

Artificial intelligence techniques have been widely used in mechatronics and robotics, especially in the emerging micro-robotics field. Meta-heuristic algorithms play a vital role in this regard. The application of an enhanced Ant Colony algorithm to a simultaneous localization and mapping

(SLAM) project for a class of mobile robots was studied in [11]. In [12], a comprehensive study on designing a pick and place robotic system has been evaluated and explored. They concluded that the ACO algorithm delivers superior performance than other meta-heuristic methods such as genetic algorithm (GA) and particle swarm optimization (PSO).

Castillo and Melin in [13] deployed particle swarm, ant colony, and genetic algorithm to enhance optimization for a fuzzy type-2 controller. The main disadvantage of using the meta-heuristic algorithm is that they propose offline optimization. If any disturbance or parameter variation occurs in the system or the environment, meta-heuristic algorithms cannot handle the system's performance. Therefore, in this article, we have presented a new controller for the system mentioned above. In [14], mission planning has been carried out for an LX robotic system using co-evolutionary optimization techniques.

In [15], a miniature modular inchworm robotic system with new mechanisms for locomotion in a micro-robotic field has been designed and evaluated experimentally. In [16], a new type of controller has been proposed for industrial Macro-Micro manipulation. A new multi-priority controller framework has been designed to use a virtual dynamic to reduce task duration and enhance the actuator's performance.

Many types of controllers have been studied and designed by researchers. In [17], a conventional PID controller was developed to control the position of a flexible robotic manipulator. PID controllers have a simple structure and easy engineering implementation. The most crucial problem is when the environments change, or uncertainty in system parameters emerges.

In [18],[19],[20], different types of sliding mode controllers have been designed to guarantee the robustness and performance of robotic systems. The fatal drawback of sliding mode controllers is chattering, which might result in instability. In [21],[22],[23], different methods of robust control techniques have been implemented for robotic manipulators. However, good stability comes at sacrificing system performance. In [24], a predictive model controller has been developed for a flexible joint manipulator. In another paper, an adaptive model predictive controller was designed for a flexible joint manipulator [25]. The primary disadvantage of model predictive controllers and their various extensions is that they necessitate a thorough understanding of the systems and environments, which is frequently impossible in real-world applications.

In [26],[27],[28],[29], different types of fuzzy controllers were applied to different robotic systems. Fuzzy controllers can provide good maneuverability and performance for robotic systems. However, they require extensive plant knowledge, which is sometimes unavailable for designers and operators.

Reinforcement learning is another popular approach for controlling complex and nonlinear systems. RL algorithms can provide good maneuverability and performance for robotic systems and deal with uncertainty in environments [30],[31],[32]. However, reinforcement learning algorithms are computationally expensive and cannot be applied to many robotic systems in real-world applications. Adaptive control is another strategy for achieving the desired behavior for robotic applications.

Generally, adaptive control strategies can be divided into two categories: classic approach and intelligent approach. Classic approach applications have been successfully applied on many robotics tasks in [33],[34],[35],[36]. However, there are several issues with classic adaptive controllers. They employ sophisticated algorithms to identify and control systems. Complex engineering implementation is another crucial problem.

On the other hand, intelligent methods are less complex in comparison with classic methods. The goal of developing AI-based adaptive control is similar to that of traditional adaptive strategies. However, they are distinct in one respect. AI-based methods can deal with a more significant amount of uncertainty than classic methods.

The aim of involving machine learning algorithms in control applications is to estimate uncertainties relatively better than pure mathematical methods. Machine learning-based methods can be widely applied to various problems and applications. Hence, we decided to select an AI-based approach to overcome the issues mentioned above. Due to the highly nonlinear nature of our plant, we are unable to use linear methods such as conventional PID controllers or offline optimal control strategies such as Genetic-PID.

Using meta-heuristic algorithms may limit the performance of controllers due to uncertainty that meta-heuristic algorithms cannot handle in the online mode. Therefore, we decided to fuse important aspects of PID controllers and machine learning methods. The advantage of PID controllers is that they are general-purpose controllers that do not require additional plant information. The key point of selecting this strategy is keeping stability and online estimation as well as adaptive behavior.

Our controller has a simple structure, unlike classic approaches for designing nonlinear adaptive controllers. The main advantage of neural networks is that they can deal with uncertainties encountered in online applications. As a result, we proposed a method for tuning the PID coefficients online and provided a linearized estimation of the plant using a neural identifier to assist the PID in updating its parameters.

This paper is organized as follows: section two discusses the theoretical aspects of the problem, such as the application of AI strategies in robotics, the AI-based PID structure, and the plant's dynamic equations. Section three details the simulation procedure and provides additional information about the plant's parametric variations. Section four discusses the results and provides critical details about the closed-loop system created using the MATLAB/Simulink software. Finally, section five summarizes the paper's major points.

## 2- Theoretical Aspects

### 2-1- Application of artificial intelligence in robotics and micro-robotics

Artificial intelligence has a variety of applications in robotics and micro-robotics, as follows:

1- Optimization to generate the robot's movement path using a dynamic model and the robot's environmental conditions

2- Neural Network applications in improving the behavioral model of robots, especially humanoid robots

3- Optimization in the field of robot assembly and structure

4- Combining optimization and path planning methods

5- Combining optimization and machine learning methods

6- Online motion control using real-time model identification

7- Developing an appropriate dynamic model using offline and online AI techniques

8- Learning and improving the model during system identification

9- Optimal inverse control methods to identify the objective function

10- Optimal control and reinforcement learning application

*2-2- Intelligent-Adaptive PID Controller with System Identification*

Today, PID controllers are widely used in industries due to their simplicity in design and high capabilities. In most cases, the adjustment of controller coefficients is adjusted by experts. When a process is delayed or slow, setting the PID controller coefficients is extremely difficult and time-consuming. If, after adjusting the coefficients, the system is affected by nonlinear factors or other factors such as perturbation, the closed-loop system's performance can change. These issues result in the need to reset the PID parameters as a result of system parameter changes.

Within the framework of the system examined in this article, we need to develop a controller that can maintain target tracking stability in the face of changing external conditions. An effective and efficient way to achieve this goal is to combine PID controllers with intelligent algorithms to improve performance and create adaptivity. These methods can receive system input and output data online and use it first to identify the system dynamics and then, based on the results, adjust and update the controller parameters. This control method is called the self-tuning PID algorithm.

In the self-tuning controller, the parameters (coefficients) are adjusted based on the automatic analysis of the process under control. Analyzing and examining the behavior of the process based on the available data from that process or the information obtained from a skilled operator leads to the creation of an approximated model of the system.

In order to control the plant in this study, two blocks of neural networks are used. Parallel to the system, the neural identifier network is designed to identify the controlled plant and calculate the controlled system's Jacobian. The second neural network is intended to adjust the PID controller's coefficients and the system output just before:

$$y_p(k-1) \tag{1}$$

the system error $(e_c(k-1) = y_d(k-1) - y_p(k-1)$ \hfill (2)

and the desired output $y_d(k-1)$. \hfill (3)

Is used as input. The main goal is to minimize error:

$$e_c(k) = y_d(k) - y_p(k) \tag{4}$$

The output of this neural network is expressed as the number of PID controller coefficients. The number of output neurons is equal to 3.

The control signal is calculated as follows:

$$u(k) = k_p(k)e_p(k) + ki(k)ei(k) + k_d(k)e_d(k) \tag{5}$$

We use a multi-layer perceptron neural network that adjusts weights using the backpropagation method and the gradient descents.

The neural network inputs are as follows:

The *x1* input is equivalent to the error between the reference value and the system's output

The *x2* input is equivalent to the error integral between the reference value and the system output

The *x3* input is equivalent to the error derivative between the reference value and the system output. Figure (1) depicts the designed neural network

The outputs of the neural network are as follows:

*Kp* output: The proportionality ratio of the PID controller

*Ki* output: The coefficient of an error-integral part corresponding to PID controller

*Kd* output: The coefficient of the part related to the error derivative in the PID controller

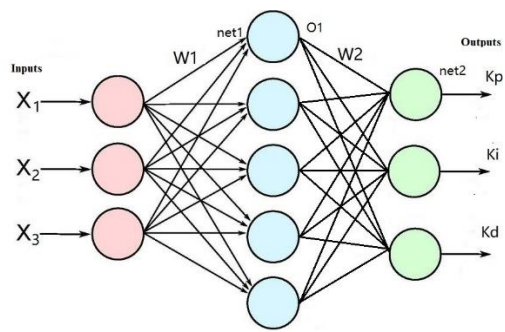

**Figure (1):** Neural Network Architecture for intelligent self-tuning algorithm

The weights of the first layer are called *W1*, and the second layer weights are called *W2*. We consider the two activation functions as a bipolar sigmoid function $(\frac{1-e^{-x}}{1+e^{-x}})$.

The output of the first layer is multiplied by the input matrix in the weights of the first layer as:

$$net1 = W_1^T u \qquad (6)$$

where we consider $u$ as the vector of neural network inputs. It should be noted that $net1$ is an input for the first activation function.

$$o^0 = f1(net1).o^0 \qquad (7)$$

Equation (7) is the output of the first activation function. Then, the input of the second layer (which is the output of the first layer) is multiplied by the weight matrix of the second layer.

$$net2 = w2^T u . \qquad (8)$$

Here, the output vector of the first layer is $(o^0)$, where this value subsequently enters the second activation function as equation (9).

$$o1 = f2(net2) \qquad (9)$$

It is necessary to explain that the number of rows in the *w1* matrix equals the number of neural network's primary inputs, while the number of columns equals the number of middle layer neurons. Similarly, *w2* is a matrix whose number of columns is equal to the number of neural network outputs (in this case, *3*), and the number of rows is equal to the number of middle layer neurons. The learning rate is *0.01* and is determined by trial and error. To adjust the weights, we consider the quadratic cost function variable as equation (10). The goal is to minimize the squared tracking error.

$$E(k) = \tfrac{1}{2}e_c^2(k) = \tfrac{1}{2}(y_d(k) - y_p(k))^2 \tag{10}$$

In equation (10), the reference and system output are at the $k_{th}$ moment. Now, in order to adjust the weights using the gradient descent, we must use the following rules:

$$\Delta w^2(k) = -\eta \frac{\partial E(k)}{\partial w^2(k)} \tag{11}$$

Using the chain derivative law, we will have a chain of weights to update weights, respectively.

$$(\Delta w^2(k) = -\eta \frac{\partial E(k)}{\partial e_c} \frac{\partial e_c}{\partial y_p} \frac{\partial y_p}{\partial u} \frac{\partial u}{\partial w^2}(k) = \eta e_c(k) J_p(k) o^1(k) \tag{12}$$

$$\Delta w^2(k) = \eta \delta^2(k) o^1(k) \tag{13}$$

$$\Delta w^1(k) = -\eta \frac{\partial E(k)}{\partial w^1(k)} = \eta e_c(k) J_p(k) w^2(k) f^{1\prime}(k) o^0(k) \tag{14}$$

$$\Delta w^1(k) = \eta \delta^1(k) o^0(k) \tag{15}$$

Now that the rules for regulating the weights of the multi-layer perceptron neural network have been obtained, we need to explain another section so that the neural network controller has complete access to the system model. In the first law, $J_p(k)$ is equivalent to the following equation.

$$J_p(k) = \frac{\partial y_p(k)}{\partial u(k_p, k_i, k_d)} \cong J_m(k) \tag{16}$$

Which indicates the Jacobian of the controlled system or plant, the values of which the neuro-identifier must immediately provide to the controller. It is sufficient to store the inputs and outputs of several plants to estimate it. The main advantage of the designed controller is the independence of the controller from the plant equations and the ability to apply the controller over any type of system. The controller, coupled with a neural system identifier, can quickly estimate the controlled plant and perform its tasks with sensors' aid. Finally, the intelligent method that controls the plant adaptively is summarized as the following block diagram.

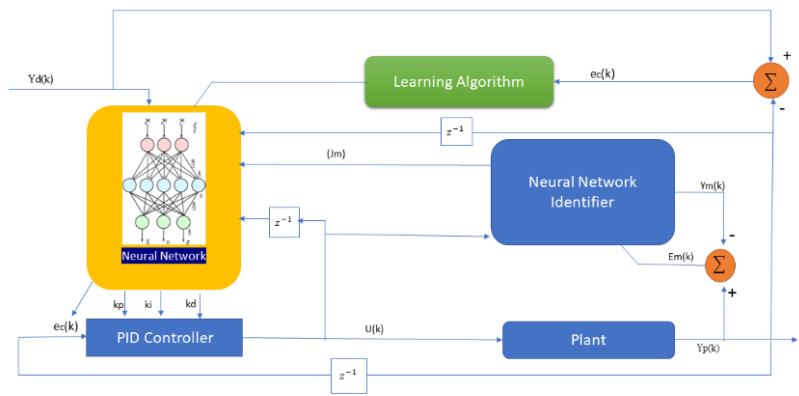

**Figure (2):** Architecture of AI-based PID Controller

## 2-3- Micro-robot dynamics equations

Micro-robot dynamics equations are derived from [37], in which they have validated these equations based on experimental investigation. In this section, we cover the most important equations for this system and its actuators. For more information, please refer to [37] to obtain more details. Proposed equations of this system are as follows:

$$M\ddot{X} = (-m_e r_e \omega^2{}_e \sin\theta_e + m_d r_d \omega^2{}_d \sin\theta_d + f_{ax} + f_{bx} + f_{cx})\cos\varphi - (f_{ay} + f_{by} + f_{cy})\sin\varphi \qquad (17)$$

$$M\ddot{Y} = (-m_e r_e \omega^2{}_e \sin\theta_e + m_d r_d \omega^2{}_d \sin\theta_d + f_{ax} + f_{bx} + f_{cx})\sin\varphi + (f_{ay} + f_{by} + f_{cy})\cos\varphi \qquad (18)$$

$$I_{zz}\ddot{\varphi} = \frac{\sqrt{3}l}{6}f_{ay} - \frac{1}{2}f_{ax} - \frac{\sqrt{3}l}{3}f_{by} + \frac{\sqrt{3}l}{6}f_{cy} + \frac{1}{2}f_{cx} - d_1 m_e r_e \omega^2{}_e \sin\theta_e - d_1 m_d r_d \omega^2{}_d \sin\theta_d \qquad (19)$$

Denoted forces to the legs in the x-direction are denoted as $f_{ax}, f_{bx}, f_{cx}$, respectively. Similarly, forces, which are applied to legs in the y-direction are denoted as $f_{ay}, f_{by}, f_{cy}$. The vibrating actuator dynamical equation is as follows:

$$\begin{cases} \omega = 2.6V^5{}_1 - 26.44V_1{}^4 + 102.11V_1{}^3 - 200.92V_1^2 + 253.09V_1 - 43.36V1 \geq 0.2 \\ \omega = 2.6V^5{}_1 + 26.44V_1{}^4 + 102.11V_1{}^3 + 200.92V_1^2 + 253.09V_1 + 43.36V1 \geq 0.2 \\ \omega = 0 - 0.2 < V1 < 0.2 \end{cases} \qquad (20)$$

In Figure (3), an open-loop diagram of the system is presented:

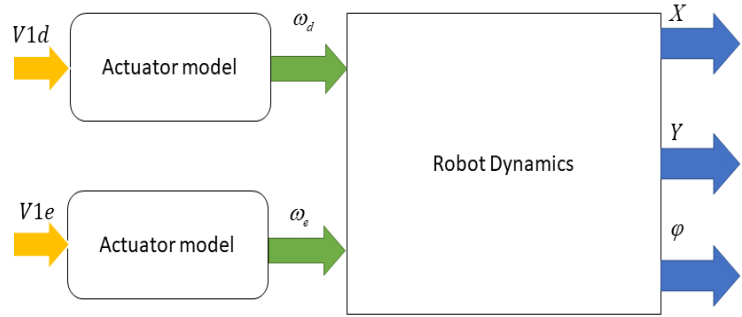

**Figure (3):** Open-loop block diagram for the micro-robot system

## 3- Micro-robot system simulation

The simulation of this micro-robot is performed in the Simulink environment of MATLAB software. This simulation includes the micro-robot dynamic equations, the friction model, and the micro-robot operator model. The simulation process can be seen in Figure (4), where the system's input is the motor voltage. Using the model chosen for the operators and the dynamics of the micro-robot, the velocity of the center of mass and the velocity of the bases are obtained. There are two methods for computing friction force, providing the velocity of the center of mass.

The spring and mass model is used if the velocity of the center of mass is zero; if it is greater than zero, the micro-class robot is considered rigid. Table (1) illustrates the values of the simulation parameters. The simulation parameters are adjusted from [37].

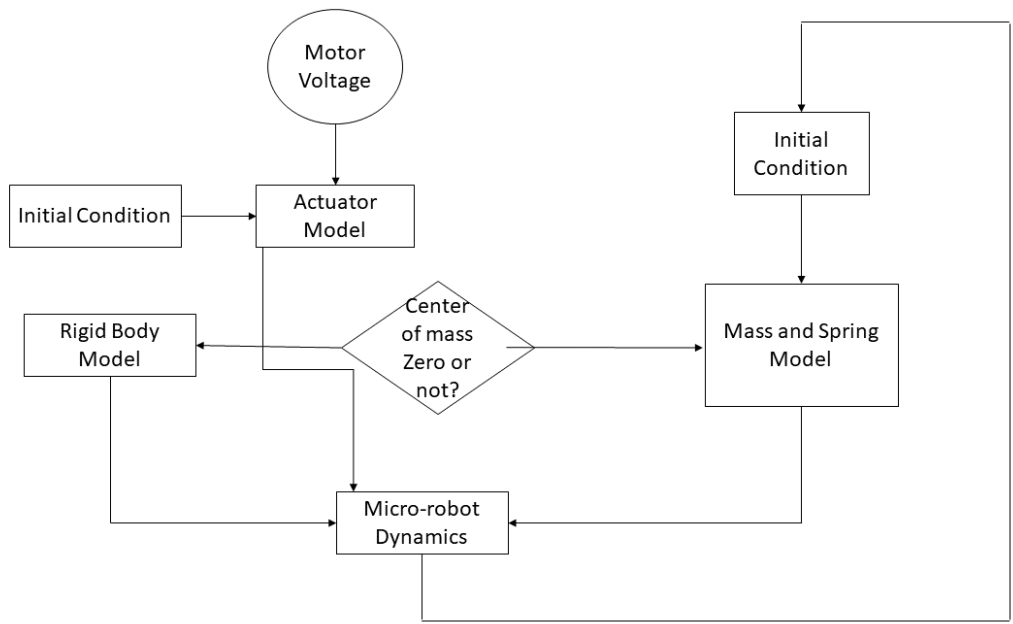

**Figure (4):** Flowchart of the simulation process

**Table (1):** Values of simulation parameters [37]

| Parameter | Value | Parameter | Value |
|---|---|---|---|
| $k$ | $72{,}509.185 (Nm^{-1})$ | $m_d$ | $9 \times 10^{-4} (kg)$ |
| $\mu$ | $0.36$ | $m_e$ | $9 \times 10^{-4} (kg)$ |
| $I$ | $9.2 \times 10^{-7} (kgm^2)$ | $r_d$ | $1.061 \times 10^{-3} (m)$ |
| $M$ | $7.2 \times 10^{-3} (kg)$ | $r_e$ | $1.061 \times 10^{-3} (m)$ |
| $l$ | $0.04 (m)$ | $R$ | $11.2 (\Omega)$ |
| d1 | 0.01(m) | L | 0.102(mH) |
| h | $5.7 \times 10^{-3}(m)$ | g | $9.807(ms^{-2})$ |

### 3-1- Investigation of the parameters of the spring coefficient of the bases and the coefficient of friction on the dynamics of the micro-robot

In this section, we examine the effect of changing the spring coefficient of the bases and the coefficient of friction on the direct motion of the micro-robot when the angle operators' velocity is 455.6 rad / s.

Figures (5) and (6) depict the translocation of the micro-class robot in the *x* and *y* directions, respectively, and Figure (7) shows the changes in the angle of the micro-robot concerning the axis perpendicular to the plane for different spring coefficients of the bases. As shown in Figure (5), the translocation of the micro-class robot in the *x-direction* has changed with increasing k. According to Figures (6) and (7), as the spring coefficient increases and the foundations harden, the displacement in the *y-direction* and the rotation around the axis perpendicular to the plane reach its ideal state.

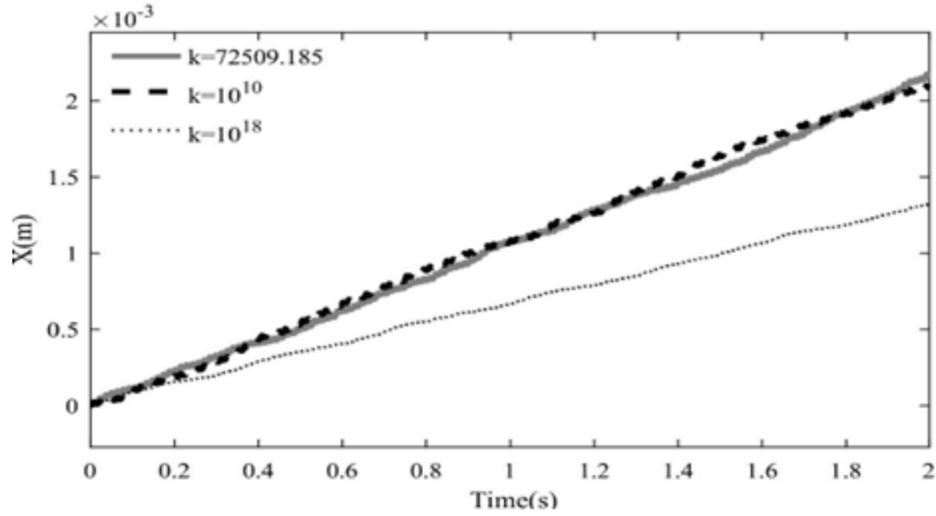

**Figure (5):** Position of the micro-class robot along the x-direction relative to time for the different spring coefficients of bases

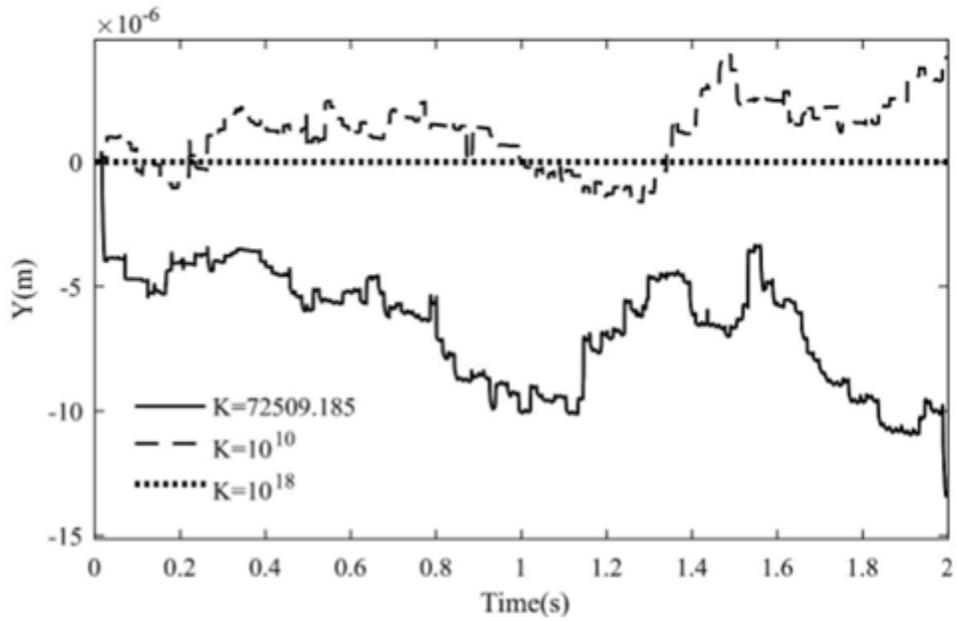

**Figure (6):** Position of the micro-class robot along the y-direction relative to time for the different spring coefficients of bases

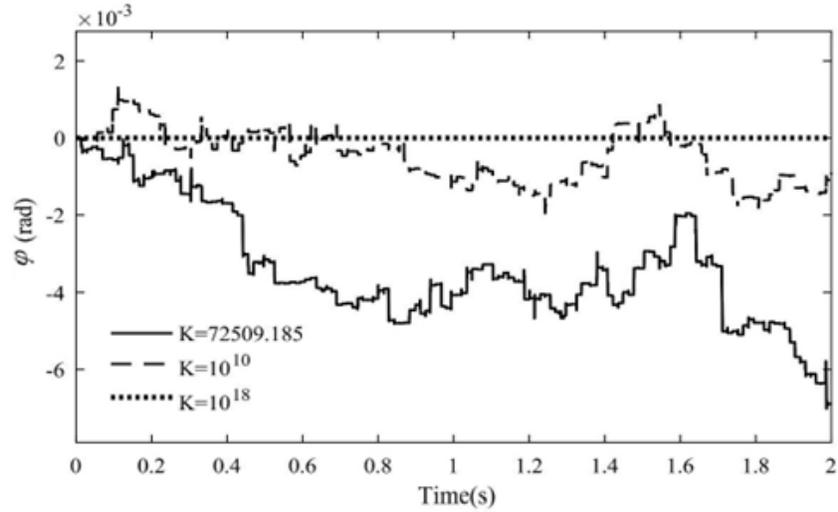

**Figure (7):** Rotation angle of the micro-class robot relative to time for the different spring coefficients of bases

Figures (8) and (9) show the translocation of the micro-class robot in the *x* and *y* directions, respectively, and Figure (10) shows the changes in its angle concerning the axis perpendicular to the surface for different coefficients of friction. As shown in Figure (8), as the coefficient of friction increases, the displacement in the *x-direction* decreases. According to Figures (9) and (10), the displacement in the *y-direction* and the rotation around the axis perpendicular to the plane is equal to zero. Because no friction exists on the surface, only the force generated by the motors in the *x-direction* enters the system, causing the micro-robot to move in this direction without deviating.

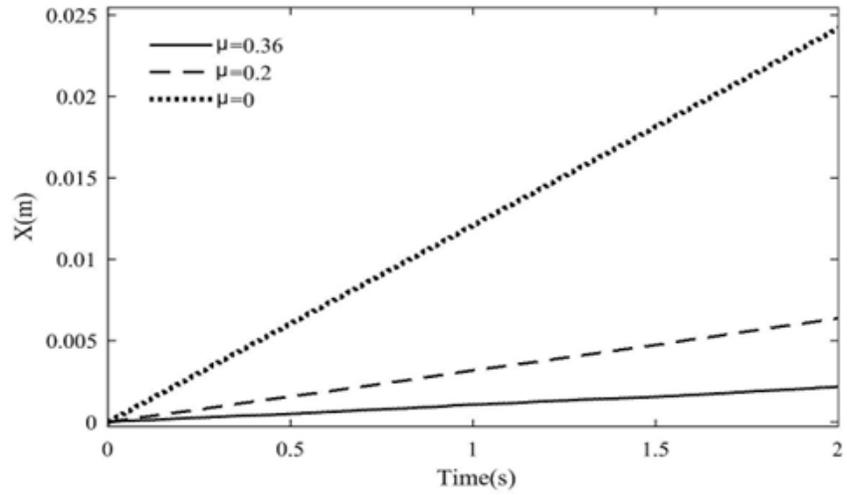

**Figure (8):** Position of the micro-class robot along the x-direction relative to time for the different coefficients of friction

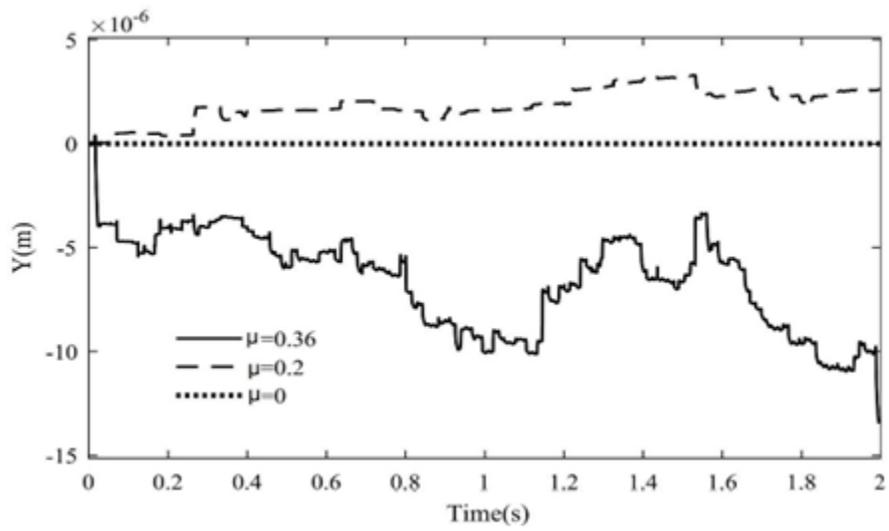

**Figure (9):** Position of the micro-class robot along the y concerning time for the different coefficients of friction

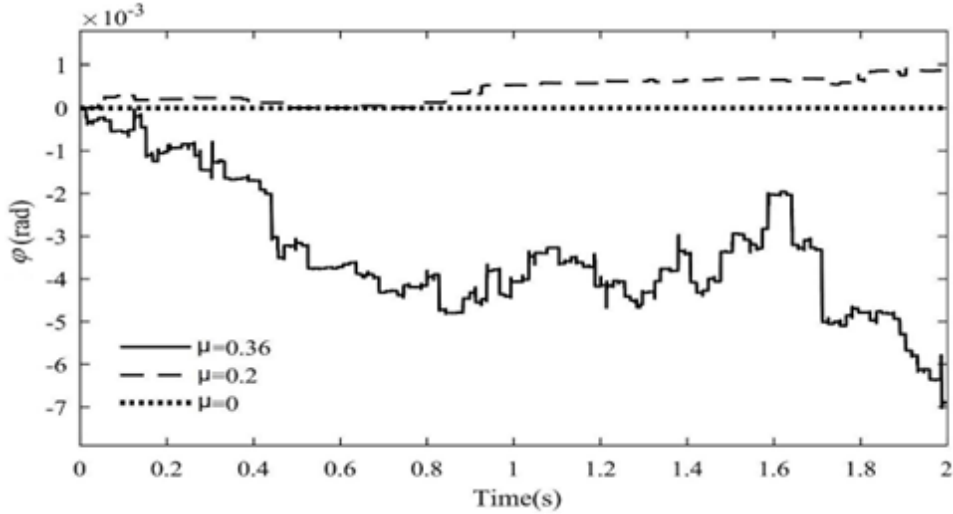

**Figure (10):** The rotation angle of the micro-robot relative to time for the different coefficients of friction

## 4-Results and Discussion

*4-1- Intelligent controller application in a micro-robot system*

In this paper, an intelligent PID controller has been used to improve robot motion control. A neural network algorithm performs the adaptation. Since this robot applies two voltages to the motors' system inputs, two Intelligent-PID controllers were used to correct the input voltage to each motor. A controller is responsible for correcting the amount of error generated by the robot's direct movement, while another controller is responsible for correcting the amount of error generated by the robot's rotation.

The critical point to remember when using the controller, particularly in practical applications such as the one above, is to maximize the controller's output. Since the controller's output in this robot is made of voltage and is applied directly to the motors, it must be defined within a permissible limit. Depending on the selected motors, this value is between ± 3v.

Figure (11) shows the block diagram of the robot with the friction force and the controllers in the Simulink environment. As shown in Figure (11), the amount of displacement in the *x-direction* concerning the x coordinate system is converted to the amount of displacement relative to the coordinate system mounted on the center of the x micro robot by the following relation. In the following equation, y is the amount of displacement in the *y-direction* relative to the reference coordinate system and $\varphi$ the amount of rotation of the microrobot relative to the reference coordinate system.

$$x = X \cos \varphi + Y \sin \varphi \tag{21}$$

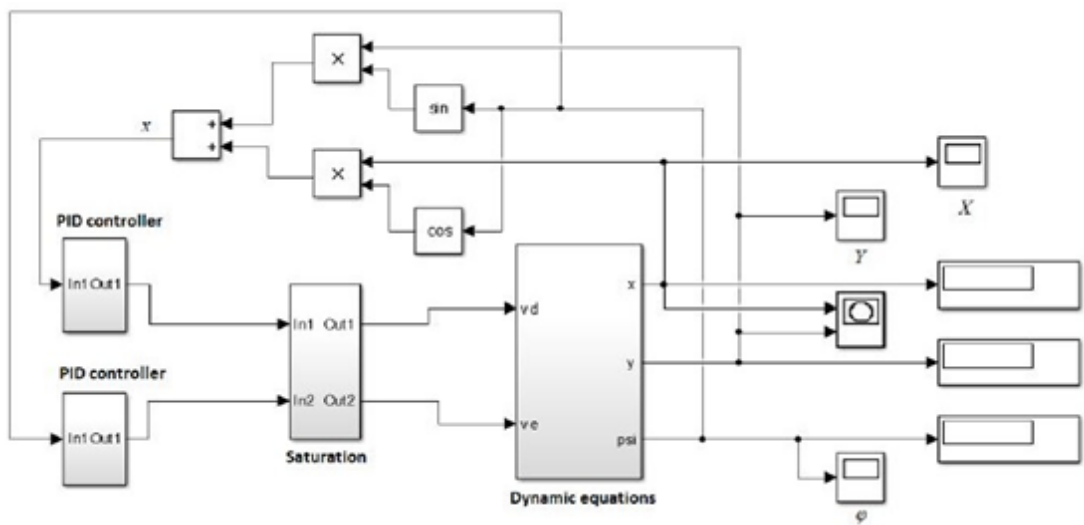

**Figure (11):** Block diagram of a closed-loop system of micro-class robot in Simulink environment

Six parameters are required to control this microrobot with the proposed PID controllers. Three parameters $K_p, K_I, K_D$ are required for direct motion, and three parameters $K_p, K_I, K_D$ for rotational motion. The aforementioned neural network algorithm selects these six parameters based on the environment and plant situation. The final target is in the form of the following equation in which:

$$e_t = x_d - x(t) \tag{22}$$

Equation (22) the error is due to direct motion, which is obtained from the difference of the reference input in the direction of displacement of the x and displacement at any moment in this direction.

$$e_r = \varphi_d - \varphi(t) \tag{23}$$

Equation (23) is the error due to net rotation, obtained from the difference between the desired rotation and the amount of rotation at each moment.

### *4-2- Micro-robot closed-loop simulation with Intelligent-Adaptive PID Controller*

The values of the micro-robot parameters used in the simulation are used from Table (1). The direct motion of the micro-robot for the following reference inputs is shown in Figures (12) to (15).

$X_d = 0.02 (m)$

$\varphi_d = 0 (rad)$

Figure (12) shows the robot moving in the *x-direction*, Figure (13) shows the amount of error related to moving in the *x-direction*, Figure (14) shows the robot moving in the *y-direction* and Figure (15) shows the rotation angle of the robot. According to Figure (13), the amount of error in the *x-direction* is minimal.

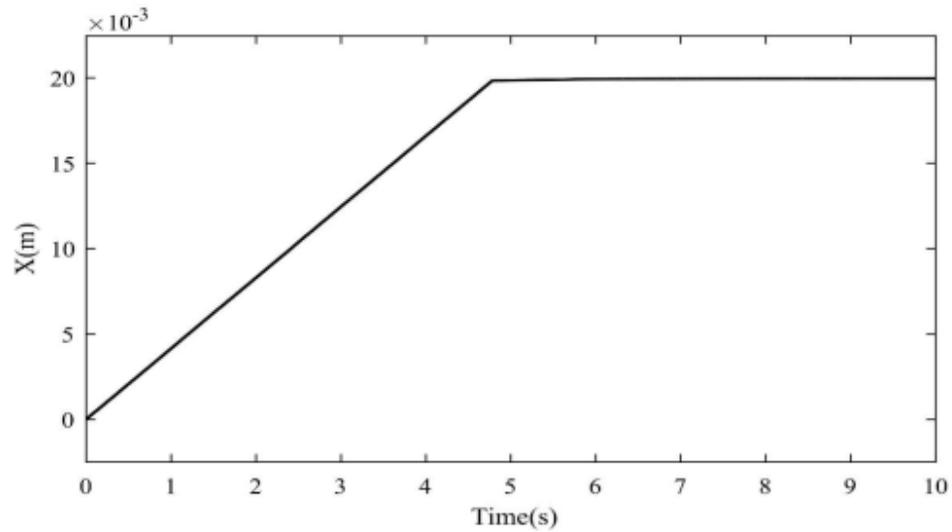

**Figure (12):** The robot motion in the x-direction

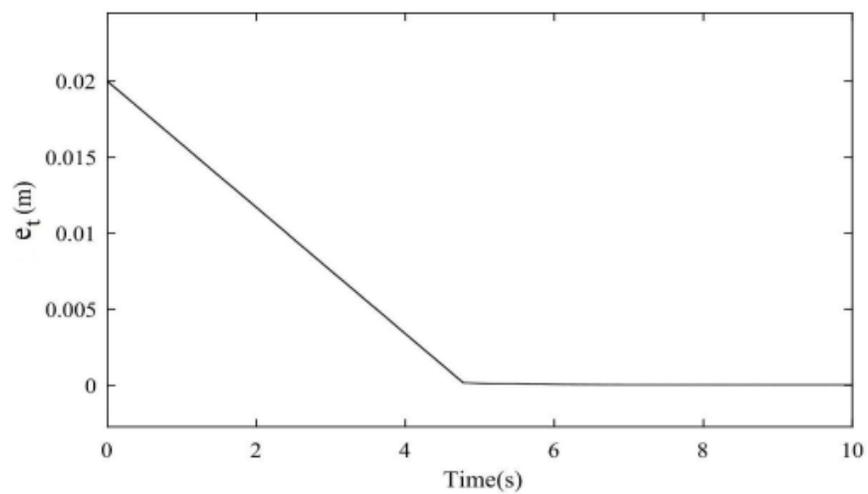

**Figure (13):** The amount of error related to displacement in the x-direction

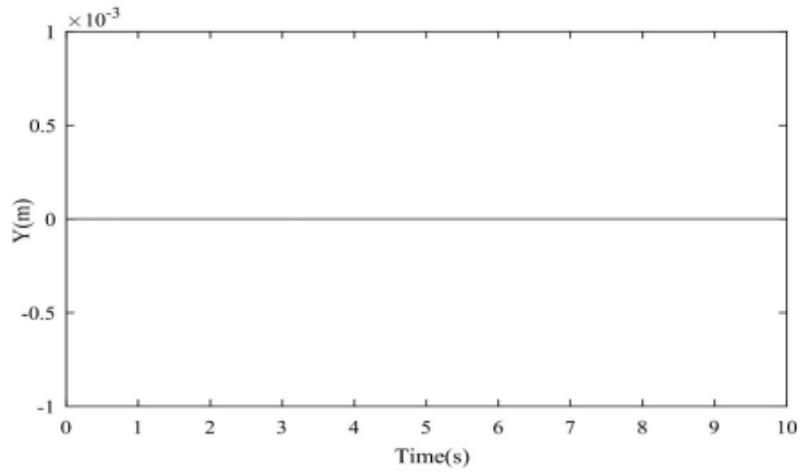

**Figure (14):** The robot motion in the y-direction

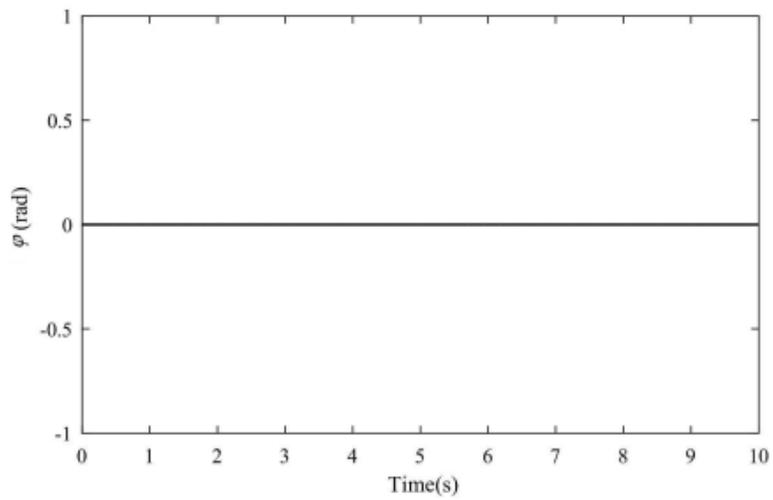

**Figure (15)**: Rotation angle of the micro-robot

The robot's rotation for the following desired inputs is shown in Figures (15) to (18).

$X_d = 0 (m)$

$\varphi_d = 0.2 (rad)$

Figure (16) shows the movement of the robot in the *x-direction*, Figure (17) shows the movement of the robot in the *y-direction*, Figure (18) shows the angle of rotation of the robot, and Figure (19) shows the amount of error related to the rotation of the robot. According to Figure (19), the amount of rotation error is minimal.

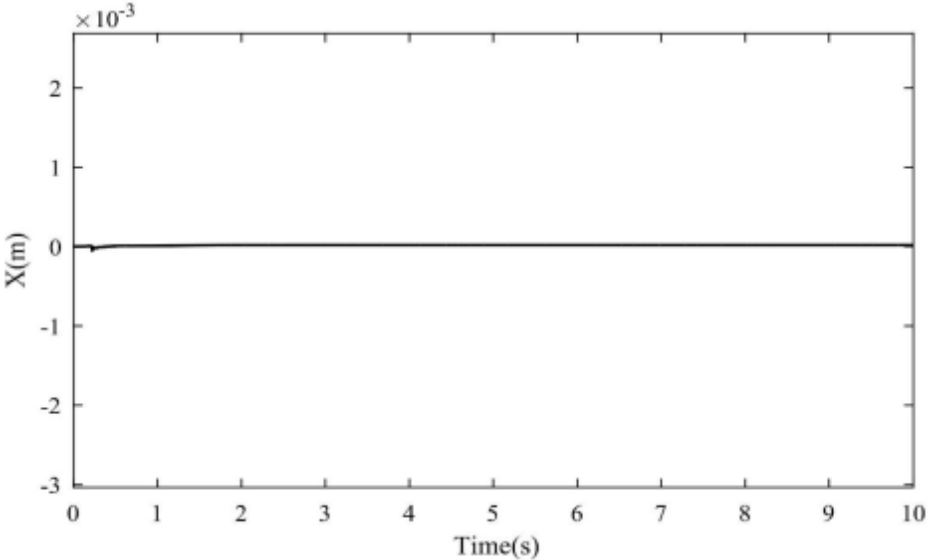

**Figure (16):** Robot motion in the x-direction

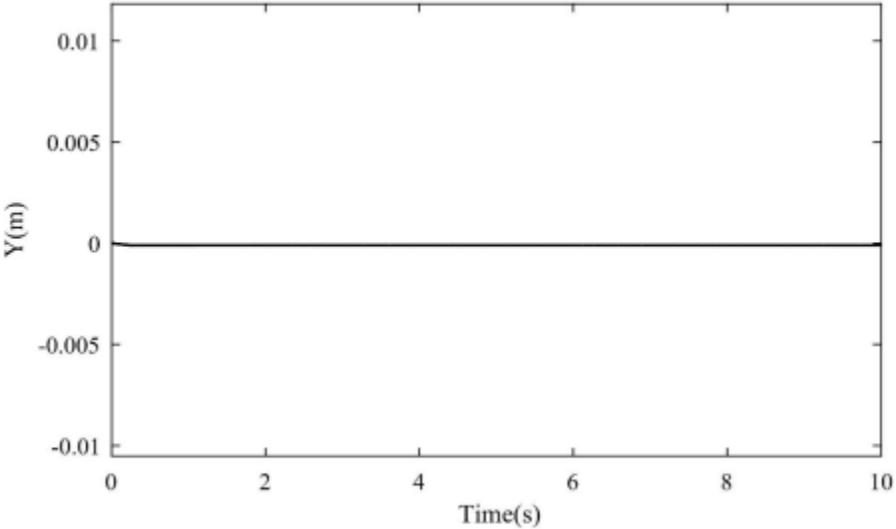

**Figure (17):** Robot motion in the y-direction

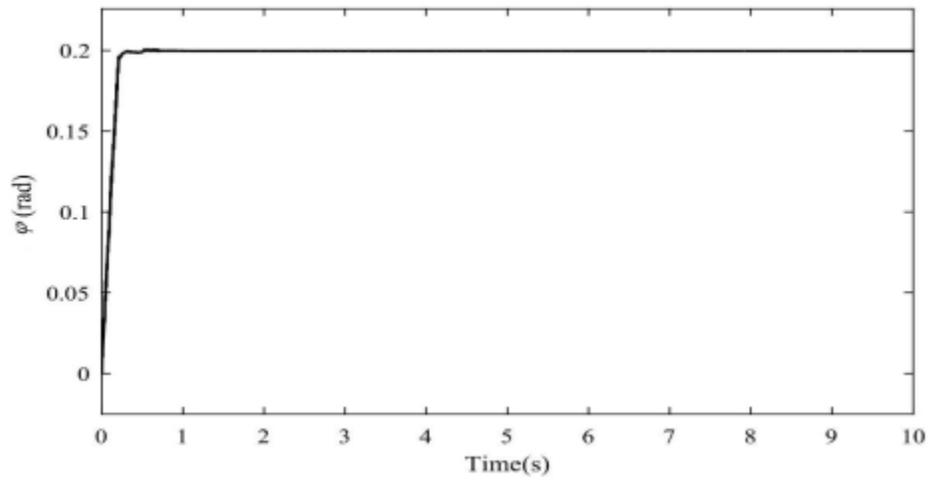

**Figure (18):** Rotation of a micro-robot

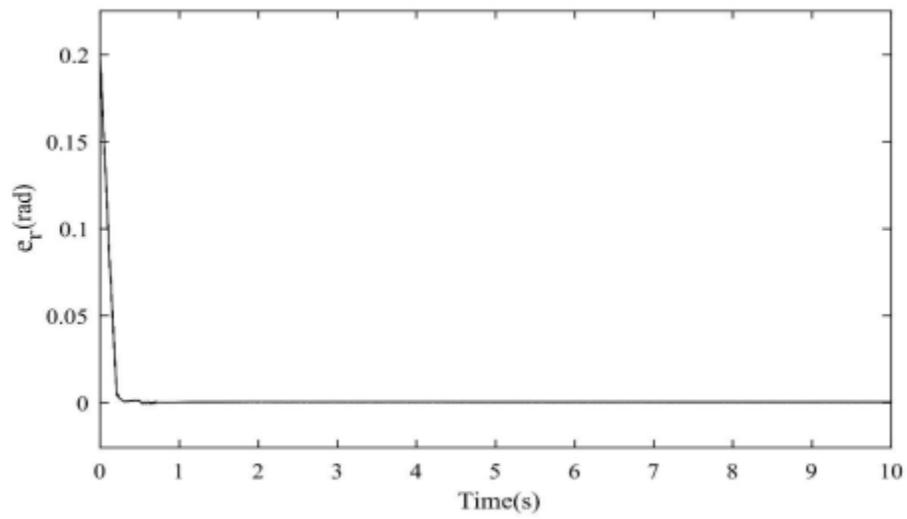

**Figure (19):** The amount of error related to the rotation of the robot

## 5- Conclusion

This paper aims to propose a controller for a micro-robot equipped with a vibration actuator. Initially, the intelligent controller has been discussed. Then, we have explained the definition of mobile micro-robots and artificial intelligence applications in robotics and micro-robotics. Our algorithm's successful performance has been demonstrated through simulations.

The micro-robot have been used in this study has followed the principle of stick-slip motion. Also, two small vibrational motors have been used to start the robot. The vibrating actuators are free of looseness and residuals and are ideal for high-precision linear and rotational movements. Dynamic modeling of this micro-robot was expressed by considering the effect of spring on the bases. Because the problem of three static indeterminate degrees needed to be solved to calculate friction forces, the problem was solved by modeling the micro-robot as a mass and spring system.

The effects of the base spring, friction coefficient, robot mass, location of actuators, and base length on the movement of the micro-robot have been investigated using MATLAB software's Simulink environment. According to the obtained results, with the increase of the spring coefficient and the hardening of the bases in direct motion, the displacement in the *y-direction* and the rotation around the axis perpendicular to the plane reached its ideal state, i.e., the value of zero. Also, by increasing the coefficient of friction in direct motion, the displacement in the *x-direction* decreases, and the displacement in the *y-direction* and the rotation around the axis vertical to the plane is equal to zero.

Because there is no friction on the surface, only the motors' energy has been transmitted to the micro-class robot in the *x-direction*, the micro-robot moves in this direction without deviating. According to simulations, the greater the mass of the micro-robot, the greater the distance between the operators and the center of mass, and the shorter the length of the bases, the less the micro-class robot translocates in a straight line, reducing its speed but increasing its accuracy of motion. In the control section, simulations for several different paths for this micro-robot movement have been performed first using the intelligent PID control method. Following that, the Intelligent-Adaptive PID controller was described, and its application was examined. The simulation results indicate an acceptable performance of the robotic system in tracking tasks.

## Author Contributions

*The authors contributed equally to this work including: Designing the study, software and code as well as writing the article.*

## Financial Support

*This research received no specific grant from any funding agency, commercial or not-for-profit sectors*

## Declaration of competing interest

*The authors declare that they have no known competing financial interests or personal relationships that could have appeared to influence the work reported in this paper.*